%% file: main.tex
\documentclass[letterpaper, 10 pt, conference]{ieeeconf}

\DeclareMathAlphabet{\altmathcal}{OMS}{cmsy}{m}{n}

\IEEEoverridecommandlockouts
\usepackage[colorlinks]{hyperref}
\usepackage[utf8x]{inputenc}
\usepackage{combelow}
\usepackage{amsmath}
\usepackage{cleveref}
\usepackage{tabularx}
\usepackage{amssymb}
\usepackage{latexsym}
\usepackage{mathptmx} % assumes new font selection scheme installed
\usepackage{subfigure}
\usepackage{flushend}
\usepackage[export]{adjustbox}

\crefname{table}{Table}{Tables}
\crefname{figure}{Figure}{Figures}
\crefname{section}{Section}{Sections}

\input{utils/custom-commands.tex}

\input{utils/acronyms.tex}
\input{utils/notation.tex}
\input{tables/table-setup.tex}

\definecolor{urlblue}{rgb}{0.1,0.1,0.5}
\hypersetup{
    colorlinks=true,    % false: boxed links; true: colored links
    linkcolor=red,      % color of internal links (change box color with linkbordercolor)
    citecolor=black,    % color of links to bibliography
    urlcolor=urlblue    % color of external links
}

\renewcommand{\baselinestretch}{0.97}

\title{Learning to Correct \Threed{}~Reconstructions from Multiple Views}

\author{%
\cb{S}tefan Săftescu$^\dagger$ \and Paul Newman$^\dagger$
\thanks{$^\dagger$Oxford Robotics Institute, University of Oxford, United Kingdom; \newline 
  {\tt\small \{stefan,pnewman\}@robots.ox.ac.uk}}
}%

\begin{document}
\maketitle

\input{sections/0_abstract.tex}
% \section{Citations}

% 	Citations can be made using either \textbackslash citep\{\} or \textbackslash citet\{\}, depending from the appropriateness. To avoid the citation moving to the next line, it is often a good practice to replace the space before with a tilde (\~{}) character.
% 	Example 1: ``CoRL is the best conference ever, as discussed in~\citep{Calandra2016}.``
% 	Example 2: ``\citet{Calandra2016} proved, both theoretically and numerically, that CoRL is the best conference ever.``
	
\input{sections/1_introduction.tex}
\input{sections/2_related-work.tex}
\input{sections/3_method.tex}
\input{sections/4_experiments.tex}
\input{sections/6_conclusion.tex}

%===============================================================================

\section*{Acknowledgment}
The authors would like to acknowledge the support of the UK’s Engineering and Physical Sciences Research Council (\textsc{epsrc}) through the Centre for Doctoral Training in Autonomous Intelligent Machines and Systems (\textsc{aims}) Programme Grant EP/L015897/1.
Paul Newman is supported by EPSRC Programme Grant EP/M019918/1.

%===============================================================================

\bibliographystyle{IEEEtran}
\bibliography{references}
~ % This prevents the last line of the bibliography from not being indented (caused by 'flushend' package)

\end{document}

%% file: utils/custom-commands.tex
\usepackage{color}
\usepackage{tikz}

\definecolor{CommentPMN}{rgb}{0.0,0.7,0.0}
\definecolor{CommentSS}{rgb}{0.0,0.4,0.6}
\definecolor{CommentMG}{rgb}{0.7,0.1,0.1}
\definecolor{Blue}{rgb}{0.0,0.0,0.7}
\newcommand{\commentthis}[3]{{{\color{#1} {\textbf{#2 says: “#3”}}}}}

\newcommand{\pmnnotes}[1] {\commentthis{CommentPMN}{Paul}{#1}}
\newcommand{\ssnotes}[1]  {\commentthis{CommentSS}{Ștefan}{#1}}
\newcommand{\mgnotes}[1]  {\commentthis{CommentMG}{Matt}{#1}}
\newcommand{\ddmnotes}[1]  {\commentthis{CommentMG}{Daniele}{#1}}

\newcommand{\ssh}[1]{{\color{CommentSS} #1}}

\renewcommand{\pmnnotes}[1] {}
\renewcommand{\ssnotes}[1]  {}
\renewcommand{\mgnotes}[1]  {}
\renewcommand{\ddmnotes}[1]  {}
\renewcommand{\ssh}[1]{}

\newcommand{\liningnums}[1]{#1}
\newcommand{\Threed}{\liningnums{3D}}
\newcommand{\Twod}{\liningnums{2D}}
\newcommand{\threed}{\Threed}
\newcommand{\twod}{\Twod}

\newcommand{\borg}{\textsc{bor\textsuperscript{\raisebox{-0.3ex}{2}}g}}

\newcommand{\gpu}{\textsc{gpu}}

\newcommand{\eg}{{e.g.}}

\newcommand{\etal}{{et~al.}}

\newcommand{\citeauthor}[2]{#2 \cite{#1}}
\newcommand{\citep}[1]{\cite{#1}}
\newcommand{\citet}[2][]{#1 \cite{#2}}

\makeatletter
\newcommand*{\getlength}[1]{\strip@pt#1}
\makeatother

%% file: utils/acronyms.tex
\usepackage[nolist]{acronym}

\begin{acronym}
  \acro{borg}[\borg{}]{Building Optimal Regularised Reconstructions with \gpu{}s}
  \acro{map}{\emph{maximum a posteriori}}
  \acro{vo}{Visual Odometry}
  \acro{imu}{Inertial Measurement Unit}
  \acro{slam}{Simultaneous Localisation and Mapping}
  \acro{sfm}[\textsc{s\textrm{f}m}]{Structure from Motion}
  \acro{ori}{Oxford Robotics Institute}
  \acro{glsl}{OpenGL Shading Language}
  \acro{mae}{mean absolute error}
  \acro{rmse}{root mean square error}
  \acro{cnn}{convolutional neural network}
  \acro{kitti-vo}{\textsc{kitti} visual odometry}
  \acro{jbf}{joint bilateral filter}
  \acro{tv}{Total Variation}
  \acro{tgv}{Total Generalised Variation}
  \acro{mlp}{multi-layer perceptron}
  \acro{dfn}{dynamic filter network}
  
\end{acronym}

%% file: utils/notation.tex
\let\outsym\Delta

\def\idepthsym{d}

\newcommand{\pred}{\ensuremath{\outsym^*}}
\newcommand{\gtruth}{\ensuremath{\outsym}}

\newcommand{\idepth}{\ensuremath{\idepthsym}}
\newcommand{\predidepth}{\ensuremath{\idepthsym^*}}
\newcommand{\lqidepth}{\ensuremath{\idepthsym^{lq}}}
\newcommand{\hqidepth}{\ensuremath{\idepthsym^{hq}}}
\newcommand{\warp}[3]{\ensuremath{#1_{#2\leftarrow{}#3}}}
\newcommand{\warpreproj}[3]{\ensuremath{\widetilde{#1_{#2;#3}}}}

\newcommand{\loss}[1]{\ensuremath{\altmathcal{L}^{#1}}}
\newcommand{\scaling}[1]{\ensuremath{\lambda^{#1}}}
\newcommand{\transform}[2]{\ensuremath{\mathbf{T}_{{#1},{#2}}}}

\def\dataname{data}

\def\gradname{\ensuremath{\nabla}}
\def\gcname{gc}
\def\regname{reg}

%% file: tables/table-setup.tex
\usepackage{pgfplotstable}
\usepackage{booktabs}

\pgfplotsset{compat=1.14}
\pgfplotstableset{
precision=2,
zerofill,
every head row/.style={before row=\toprule,after row=\midrule},
every last row/.style={after row=\bottomrule},
col sep=comma,
trim cells=true,
string type,
assign column name/.style={
    /pgfplots/table/column name={\textbf{#1}}
},
columns/imae/.style={ column name=i\textsc{mae}, sci, numeric type },
columns/irmse/.style={ column name=i\textsc{rmse}, sci, numeric type },
columns/mae/.style={ column name=\textsc{mae}, sci, numeric type },
columns/rmse/.style={ column name=\textsc{rmse}, sci, numeric type },
columns/thacc1.05/.style={ preproc/expr={100*##1} },
columns/thacc1.10/.style={ preproc/expr={100*##1} },
columns/thacc1.15/.style={ preproc/expr={100*##1} },
columns/thacc1.25/.style={ preproc/expr={100*##1} },
columns/thacc1.56/.style={ preproc/expr={100*##1}},
columns/thacc1.95/.style={ preproc/expr={100*##1} },
columns/delta1/.style={ preproc/expr={100*##1} },
columns/delta2/.style={ preproc/expr={100*##1} },
columns/delta3/.style={ preproc/expr={100*##1} },
thacc column/.style={%
    /pgfplots/table/columns/thacc#1/.append style={%
        column name=$\mathbf{\delta < {#1}}$, 
        numeric type, 
        fixed,
        fixed zerofill,
        postproc cell content/.append code={
            \pgfkeysalso{@cell content/.add={}{\%}}%
        },
    } 
},
delta column/.style={%
    /pgfplots/table/columns/delta#1/.append style={%
        column name=$\mathbf{\delta} < {1.25^#1}$, 
        numeric type, 
        fixed,
        fixed zerofill,
        dec sep align,
        postproc cell content/.append code={
            \ifnum1=\pgfplotstablepartno
                \pgfkeysalso{@cell content/.add={}{\%}}%
            \fi
        }
    } 
},
delta column/.list={1,2,3},
thacc column/.list={1.05,1.10,1.15,1.25,1.56,1.95},
}

\newcommand{\findmax}[3]{
    \pgfplotstablevertcat{\datatable}{#1}
    \pgfplotstablecreatecol[
    create col/expr={%
    \pgfplotstablerow
    }]{rownumber}\datatable
    \pgfplotstablesort[sort key={#2},sort cmp={float >}]{\sorted}{\datatable}%
    \pgfplotstablegetelem{0}{rownumber}\of{\sorted}%
    \pgfmathtruncatemacro#3{\pgfplotsretval}
    \pgfplotstableclear{\datatable}
}

\newcommand{\findsecondmax}[3]{
    \pgfplotstablevertcat{\datatable}{#1}
    \pgfplotstablecreatecol[
      create col/expr={%
    \pgfplotstablerow
    }]{rownumber}\datatable
    \pgfplotstablesort[sort key={#2},sort cmp={float >}]{\sorted}{\datatable}%
    \pgfplotstablegetelem{1}{rownumber}\of{\sorted}%
    \pgfmathtruncatemacro#3{\pgfplotsretval}
    \pgfplotstableclear{\datatable}
}

\newcommand{\findmin}[3]{
    \pgfplotstablevertcat{\datatable}{#1}
    \pgfplotstablecreatecol[
    create col/expr={%
    \pgfplotstablerow
    }]{rownumber}\datatable
    \pgfplotstablesort[sort key={#2},sort cmp={float <}]{\sorted}{\datatable}%
    \pgfplotstablegetelem{0}{rownumber}\of{\sorted}%
    \pgfmathtruncatemacro#3{\pgfplotsretval}
    \pgfplotstableclear{\datatable}
}

\newcommand{\findsecondmin}[3]{
    \pgfplotstablevertcat{\datatable}{#1}
    \pgfplotstablecreatecol[
      create col/expr={%
    \pgfplotstablerow
    }]{rownumber}\datatable
    \pgfplotstablesort[sort key={#2},sort cmp={float <}]{\sorted}{\datatable}%
    \pgfplotstablegetelem{1}{rownumber}\of{\sorted}%
    \pgfmathtruncatemacro#3{\pgfplotsretval}
    \pgfplotstableclear{\datatable}
}

\pgfplotstableset{
    highlight col max/.code 2 args={
        \findmax{#1}{#2}{\maxval}
        \edef\setstyles{\noexpand\pgfplotstableset{
                every row \maxval\noexpand\space column #2/.style={
                    postproc cell content/.append style={
                        /pgfplots/table/@cell content/.add={$\noexpand\bf}{$}
                    },
                }
            }
        }\setstyles
    },
    highlight col secondmax/.code 2 args={
        \findsecondmax{#1}{#2}{\maxval}
        \edef\setstyles{\noexpand\pgfplotstableset{
                every row \maxval\noexpand\space column #2/.style={
                    postproc cell content/.append style={
                        /pgfplots/table/@cell content/.add={$\noexpand\it}{$}
                    },
                }
            }
        }\setstyles
    },
    highlight col min/.code 2 args={
        \findmin{#1}{#2}{\minval}
        \edef\setstyles{\noexpand\pgfplotstableset{
                every row \minval\noexpand\space column #2/.style={
                    postproc cell content/.append style={
                        /pgfplots/table/@cell content/.add={$\noexpand\bf}{$}
                    },
                }
            }
        }\setstyles
    },
    highlight col secondmin/.code 2 args={
        \findsecondmin{#1}{#2}{\minval}
        \edef\setstyles{\noexpand\pgfplotstableset{
                every row \minval\noexpand\space column #2/.style={
                    postproc cell content/.append style={
                        /pgfplots/table/@cell content/.add={$\noexpand\it}{$}
                    },
                }
            }
        }\setstyles
    },
}

\makeatletter
\long\def\pgfplotstabletypeset@opt@collectarg[#1]#2{%

    \pgfplotstable@isloadedtable{#2}%
        {\pgfplotstabletypeset@opt@[#1]{#2}}%
        {\pgfplotstabletypesetfile@opt@[#1]{#2}}%
}
\makeatother

%% file: sections/0_abstract.tex
\begin{abstract}
This paper is about reducing the cost of building good large-scale \threed{} reconstructions post-hoc.
We render \twod{} views of an existing reconstruction and train a \ac{cnn} that refines inverse-depth to match a higher-quality reconstruction.
Since the views that we correct are rendered from the same reconstruction, they share the same geometry, so overlapping views complement each other.
We take advantage of that in two ways.
Firstly, we impose a loss during training which guides predictions on neighbouring views to have the same geometry and has been shown to improve performance.
Secondly, in contrast to previous work, which corrects each view independently, we also make predictions on sets of neighbouring views jointly.
This is achieved by warping feature maps between views and thus bypassing memory-intensive \threed{} computation.
We make the observation that features in the feature maps are viewpoint-dependent, and propose a method for transforming features with dynamic filters generated by a \acl{mlp} from the relative poses between views.
In our experiments we show that this last step is necessary for successfully fusing feature maps between views.
\end{abstract}

%% file: sections/1_introduction.tex
\section{Introduction}
\label{sec:introduction}

Building good dense \threed{} reconstructions is essential in many robotics tasks, such as surveying, localisation, or planning.
Despite numerous advancements in both hardware and algorithms, large-scale reconstructions remain costly to build.

\input{figures/feature-aggregation-small.tex}
We approach this issue by trying to reduce the data acquisition cost either through the use of cheaper sensors, or by collecting less data.
To make up for the cheaper but lower quality data, we have to turn to prior information from the operational environment (\eg{} roads and buildings are flat, cars and trees have specific shapes, etc).
To learn these priors, we train a \ac{cnn} over \twod{} views of \threed{} reconstructions, and predict refined inverse-depth maps that can be fused back into a refined \threed{} reconstruction.
We take this detour through two dimensions in order to avoid the high memory requirements a volumetric approach over large-scale reconstructions would impose.

While operating in \twod{}, neighbouring views are related by the underlying geometry.
Previous work \citep{Saftescu:2019:GC} has leveraged this relation during training, where a geometric consistency loss is imposed between neighbouring views that penalises mismatched geometry. 
Here, we take another step and explore how neighbouring views can be used together when \emph{predicting} refined depth, and to that end introduce a method for aggregating feature maps in the \ac{cnn}.

To fuse feature maps from multiple views, we could either “un-project” them into a common \threed{} volume or “collect” them into a common target view through reprojection, as proposed by \cite{Donne:2019:DeFuSR}.
As un-projecting into \threed{} re-introduces the limitation we wished to avoid, we take the latter approach.

Directly aggregating feature maps between views – either in a \threed{} volume or in a common target view – implies features are somewhat independent of viewpoint.
To lift this restriction, we propose a method for transforming \emph{features} between views, enabling us to more easily aggregate feature maps from arbitrary viewpoints.
Concretely, we use the relative pose between views to generate a projection matrix in feature space that can be used to transform feature maps, as illustrated in Figure~\ref{fig:feature-aggregation}.

This paper brings the following contributions:
\begin{enumerate}
    \item We introduce a method for fusing multi-view data that decouples much of the multi-view geometry from model parameters.
    Not only do we warp feature maps between views, but we make the key observation that features \emph{themselves} can be view-point dependent, and show how to transform the \emph{feature space} between views.
    \item We apply this method to the problem of correcting dense \threed{} meshes.
    We render \twod{} views from reconstructions and learn how to refine inverse-depth, while making use of multi-view information.
\end{enumerate}

In our experiments, we look at two ways of aggregating feature maps, and conclude that the feature space transformation is necessary to benefit from the use of multiple views when correcting reconstructions.

%% file: figures/feature-aggregation-small.tex
\begin{figure}[h!]
\centering
\includegraphics[width=\columnwidth]{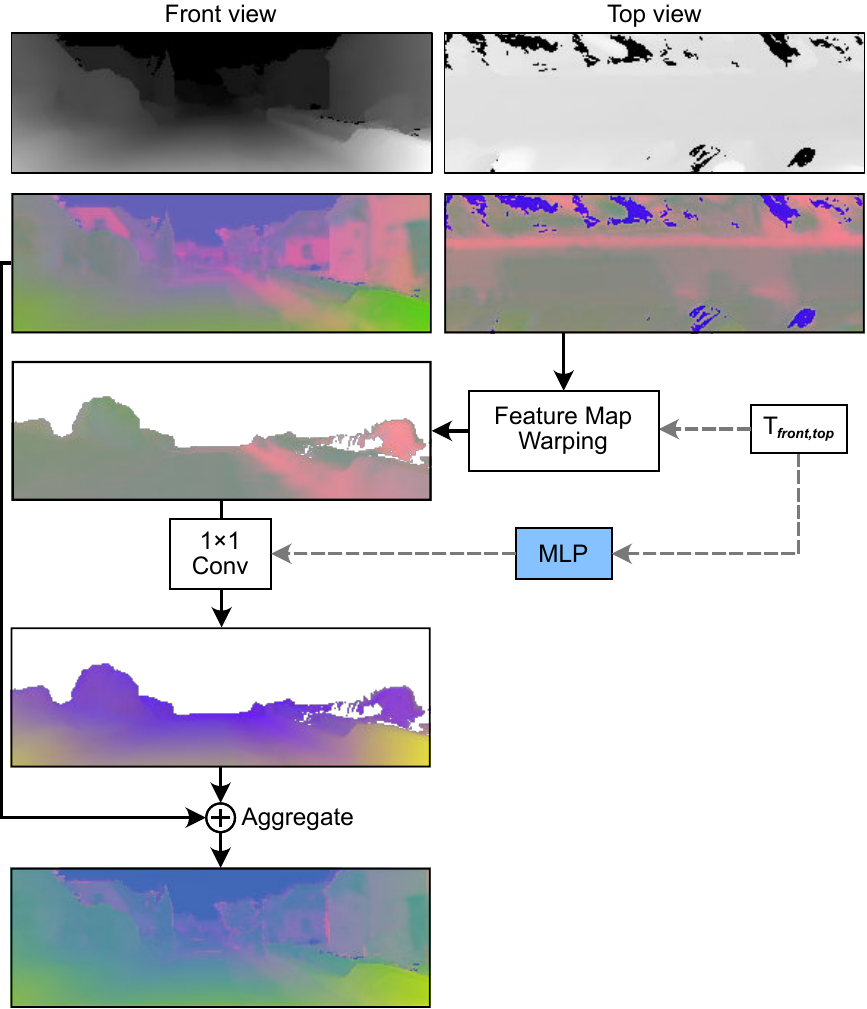}
\caption{
Feature map aggregation.
In the top two rows we show inverse-depth images for a front view and a top view, along with two feature maps.
To aggregate the feature map of the top view with the front view, we first warp the top feature map into the front view.
The relative transform between the top and the front view, $\transform{front}{top}$, is processed by a \acl{mlp} (\acs{mlp}) to generate the weights for a linear transform that maps the \emph{features} from the top view to the front view.
Finally, the resulting feature map can be aggregated with the front view feature map.
Note that in the front feature map the features fade from green to violet towards the horizon, while in the warping of the top view, the features do not change with depth.
Only after transforming the \emph{features} do we see them fading towards the horizon.
Analogously, we aggregate the front view feature map with the top view one.
For visualisation, the above feature maps are projected to three channels using the same random projection.
}
\label{fig:feature-aggregation}
\end{figure}

%% file: sections/2_related-work.tex
\section{Related Work}
\label{sec:related-work}
Our work focuses on refining the \emph{output} of an existing \threed{} reconstruction system such as \borg{} \citep{Tanner:2015:BORG} or KinectFusion \citep{Newcombe:2011:KinectFusion}, thus producing higher-quality reconstructions.
Since we achieve this by operating on \twod{} projections and refining inverse-depth, our work is related to depth refinement as well.
In the following we summarise some of the related literature and methods used in this work.

\paragraph*{Mesh correction}
\citet[Tanner \etal{}]{Tanner:2018:MeshedUp} first propose fixing \threed{} reconstructions by refining \twod{} projections of them with a \ac{cnn}, one at a time.
The geometrical relation between neighbouring views is leveraged in \cite{Saftescu:2019:GC} during training, by the addition of a geometric consistency loss that penalises differences in geometry. 
In this work, we process neighbouring views jointly not only while training, but also when making predictions.

\paragraph*{Learnt depth refinement and completion}
% There are several methods for learning depth map completion.
% For example, \cite{Uhrig:2017:SparsityInvariant,Eldesokey:2018:UncertaintySparsity} train a \ac{cnn} to predict dense depth maps from sparse laser data.
% Similarly, \cite{Hua:2018:NConvDepthCompletion} use normalised convolutions \citep{Knutsson:1993:NormalizedConv} to recover dense depth maps from sparse samples.
% These methods only fill in missing data and rely on sparse, high-quality input – either lidar scans, or ground truth samples.
% Our method only requires high-quality meshes during training.
% During inference, we predict both missing data and refine existing surfaces directly on low-quality input.

There are several depth refinement methods similar to our approach.
\cite{Kwon:2015:DataDrivenDepth} fuse multiple depth maps with KinectFusion \citep{Newcombe:2011:KinectFusion} to obtain a high-quality reference mesh, and use dictionary learning to refine raw \textsc{rgb-d} images.
Using a \ac{cnn} on the colour channels of an \textsc{rgb-d} image, \cite{Zhang:2018:DeepDepthCompletion} predict normals and occlusion boundaries and use them to optimise the depth component, filling in holes.
\cite{Jeon:2018:ReconstructionPairwiseDataset} render depth images from a reconstruction at the same locations as the raw depth, obtaining a 4000-image dataset of raw/clean depth image pairs.
The authors train a \ac{cnn} to refine the raw depth maps, and show that using it reduces the amount of data and time needed to build \threed{} reconstructions.
All these methods require a colour image in order to refine depth, and operate on live data, which limits the amount of training data available.
Our method is designed to operate post-hoc, on existing meshes.
We can therefore generate an arbitrary number of training pairs from any viewpoint, removing any viewpoint-specific bias that might otherwise surface while learning.

Another recent approach proposes depth refinement by fusing feature maps of neighbouring views through warping \cite{Donne:2019:DeFuSR}.
While this is similar to our approach, we take the additional step of transforming the \emph{features} between views, and consider two feature aggregation methods.

\paragraph*{Dynamic filter networks}
Generating filters for convolutions dynamically conditioned on network inputs is presented by \cite{Brabandere:2016:DFN}, where the method is used to predict filters for local spatial transforms that help in video prediction tasks.
Our feature transformation is also based on this framework: given a relative transform between two views, we predict the weights that would transform features from one view to another.
A key distinction is that, while the filters in the original work are demonstrated over the spatial domain, we operate solely on the channels of the feature map with a $1\times1$ convolution.

%% file: sections/3_method.tex
\section{Mehtod}
\label{sec:method}
\input{figures/mesh-features.tex}
%%%%%%%%%%%%%%%%%%%%%%%%%%%%%%%%%%%%%%%%%%%%%%%%%%%%%%%%%%%%%%%%%%%%%%%%%%%%%%%%
%%%%%%%%%%%%%%%%%%%%%%%%%%%%%%%%%%%%%%%%%%%%%%%%%%%%%%%%%%%%%%%%%%%%%%%%%%%%%%%%
\subsection{Training Data} 
Our main goal is to correct \emph{existing} dense \threed{} reconstructions.
To bypass the need for expensive \threed{} computation, we operate on \twod{} projections of a mesh from multiple viewpoints.
As we want to capture as much of the \threed{} geometry as possible in our projections, we render several mesh features for each viewpoint: inverse-depth, colour, normals, and triangle surface area (see \cref{fig:mesh-features}).

During training, we have access to two reconstructions of the same scene: a low-quality one that we learn to correct, and a high-quality one that we use for supervision.
In particular, we learn to correct stereo-camera reconstructions using lidar reconstructions as supervision.
Figure~\ref{fig:training} shows an overview of our method.

The ground-truth labels $\gtruth$ are computed as the difference in inverse-depth between high-quality and low-quality reconstructions:

\begin{equation}\label{eq:g}
  \gtruth(\mathbf{p}) = \hqidepth(\mathbf{p}) - \lqidepth(\mathbf{p})
\end{equation}%
where $\mathbf{p}$ is a pixel index, and $\hqidepth$ and $\lqidepth$ are inverse-depth images for the high-quality and low-quality reconstruction, respectively.
For notational compactness, $\gtruth(\mathbf{p})$ is referred to as $\gtruth$, and future definitions are over all values of $\mathbf{p}$, unless otherwise specified. 

There are several advantages to using inverse-depth.
Firstly, geometry closer to the camera will have higher values and therefore these areas will be emphasised during training.
Secondly, inverse-depth smoothly fades away from the camera, such that the background – that has no geometry and is infinitely far away – has a value of zero.
If we were to use depth, we would have to treat background as a special case, since neural networks are not equipped to deal with infinite values out-of-the-box.
Finally, when warping images from one viewpoint from another, as described in the following sections, we are, in essence, re-sampling.
To correctly interpolate depth values, we would have to use harmonic mean, which is less numerically stable, whereas interpolating inverse-depth can be done linearly.

%%%%%%%%%%%%%%%%%%%%%%%%%%%%%%%%%%%%%%%%%%%%%%%%%%%%%%%%%%%%%%%%%%%%%%%%%%%%%%%%
%%%%%%%%%%%%%%%%%%%%%%%%%%%%%%%%%%%%%%%%%%%%%%%%%%%%%%%%%%%%%%%%%%%%%%%%%%%%%%%%
\subsection{Image Warping}
During both training and prediction, we need to fuse information from neighbouring views.
While training, we want to penalise the network for making predictions that are geometrically inconsistent between views, using the geometric consistency loss from \cite{Saftescu:2019:GC}, described in \cref{sec:losses}.
When making predictions, we want to be able to aggregate information from multiple views.
To enable this, we need to warp images between viewpoints, such that corresponding pixels are aligned.

Consider a view $t$, an inverse-depth image $\idepth_t$, and a pixel location within the image, $\mathbf{p}_t = [u ~ v ~ 1]^T$.
The homogeneous \threed{} point corresponding to $\mathbf{p}_t$ is:
\begin{equation}
\mathbf{x}_t = \left[
\begin{matrix}
    \mathbf{K}^{-1}\mathbf{p}_t \\
    \idepth_t(\mathbf{p}_t)
\end{matrix}\right],
\end{equation}
where $\mathbf{K}$ is the camera intrinsic matrix.
Consider further a nearby view $n$, an image $I_n$, and the relative transform $\transform{n}{t} \in \mathrm{SE}(3)$ from $t$ to $n$ that maps \threed{} points between views:
$\mathbf{x}_n = \transform{n}{t}\mathbf{x}_t$.
The pixel in view $n$ corresponding to $\mathbf{p}_t$ is:
\begin{equation}
    \mathbf{p}_n = \mathbf{K} \frac{\mathbf{x}_n^{(1:3)}}{\mathbf{x}_n^{(3)}},
\end{equation}
where the superscript indexes into the vector $\mathbf{x}_n$.

We can now \emph{warp} image $I_n$ into view $t$:
\begin{equation}\label{eqn:image-warp}
    \warp{I}{t}{n}(\mathbf{p}_t) = I_n(\mathbf{p}_n).
\end{equation}
Note here that $\mathbf{p}_n$ might not have integer values, and therefore might not lie exactly on the image grid of $I_n$.
In that case, we linearly interpolate the nearest pixels.

Since the value of inverse-depth is view-dependent, when warping inverse-depth images we make the following additional definition:
\begin{equation}\label{eqn:idepth-warp}
\warpreproj{\idepth}{t}{n}(\mathbf{p}_t) = \frac{\mathbf{x}_n^{(4)}}{\mathbf{x}_n^{(3)}},
\end{equation}
which represents an image aligned with view $t$, with inverse-depth values in the frame of $n$.

%%%%%%%%%%%%%%%%%%%%%%%%%%%%%%%%%%%%%%%%%%%%%%%%%%%%%%%%%%%%%%%%%%%%%%%%%%%%%%%%
\paragraph*{Occlusions}
Pixel correspondences computed through warping are only valid where there are no occlusions.
We therefore need a mask to only take into account unoccluded regions.
When rendering mesh features, we also render an additional image where every pixel is assigned the \textsc{id} of the visible mesh triangle at that location.
The triangle \textsc{id} is computed by hashing the global coordinates of its vertices.
We can then warp this image of triangle \textsc{id}s from the source to the target view.
If the \textsc{id} of a pixel matches between the warped source and the target image, we know that the same surface is in view in both images, and thus that pixel is unoccluded.

%%%%%%%%%%%%%%%%%%%%%%%%%%%%%%%%%%%%%%%%%%%%%%%%%%%%%%%%%%%%%%%%%%%%%%%%%%%%%%%%
%%%%%%%%%%%%%%%%%%%%%%%%%%%%%%%%%%%%%%%%%%%%%%%%%%%%%%%%%%%%%%%%%%%%%%%%%%%%%%%%
\subsection{Network architecture}
\input{figures/training.tex}
\input{tables/network-architecture.tex}
\subsubsection{Model}
We use an encoder-decoder architecture, with asymmetric ResNet \cite{He:2016:ResNet} blocks, the sub-pixel convolutional layers proposed in \cite{Shi:2016:PixelShuffle} for upsampling in the decoder, and skip connections between the encoder and the decoder to improve sharpness, as introduced by the U-Net architecture \cite{Ronneberger:2015:UNet}. Throughout the network, we use \textsc{elu} \cite{Clevert:2015:ELU} activations and group normalisation \cite{Wu:2018:GroupNorm}.
Table~\ref{tab:network-architecture} details the blocks used in our network.

Since a fair portion of the input low-quality reconstruction is already correct, we train our model to predict the error in the input inverse-depth, $\pred$.
We then compute the refined inverse depth as the output of our network:

\begin{equation}
    \predidepth = \textrm{max}(\lqidepth + \pred, 0).
\end{equation}

Clipping is required here because inverse-depth cannot be negative.
However, since we are supervising the predicted error, the network can learn even when the predicted inverse-depth is clipped and would therefore lack a gradient.
To ensure our network can deal with any range of inverse-depth, we offset the input such that it has zero mean, scale it to have standard deviation of 1, and undo the scaling on the predicted error \pred.

%%%%%%%%%%%%%%%%%%%%%%%%%%%%%%%%%%%%%%%%%%%%%%%%%%%%%%%%%%%%%%%%%%%%%%%%%%%%%%%%
\subsubsection{Feature Map Warping and Aggregation}\label{sec:arch:fmap-warp}
As our predictions are related by the \threed{} geometry of a scene, we would like to ensure predictions are consistent between views.
This is taken into account during training by using the geometric consistency loss from \cite{Saftescu:2019:GC}, as described in Section~\ref{sec:losses}.

However, during inference we would like to aggregate information from multiple views to improve predictions.
Take, for example, two views, $t$ and $n$, and feature maps in the network, $F_t$ and $F_n$, after a certain number of layers, corresponding to each of the views.
We would like to aggregate them such that $F_t \oplus F_n$ is a feature map containing information from both views.
Since the feature maps are aligned with the input views, we cannot do that pixel-wise.
Using the input depth, we warp the feature map of one view into the frame of the other, such that the input geometry is aligned.
For the two input views, we can thus warp $F_n$ to the viewpoint of $t$, and then aggregate the two feature maps: $F_{t,n} = F_t \oplus \warp{F}{t}{n}$, obtaining a feature map aligned with $t$ that combines information from both views.

The aggregation step is necessary (instead of simply concatenating aligned feature maps) to allow for an arbitrary number of views.
For the same reason, the aggregation function needs to be invariant to permutations of views.
We consider two such aggregation functions: averaging, and attention.
For a target view $t$ with neighbourhood $N = \{t\} \cup \{n_1, n_2, ...\}$, averaging is defined as $\sum_{n \in N} \warp{F}{t}{n}/|N|$, where we consider $\warp{F}{t}{t} \equiv F_t$.

For the attention-based aggregation, we use the attention method proposed by \citet[Bahdanau et al.]{Bahdanau:2014:Attention}.
A per-pixel score $E_{t,n} = a(F_t, \warp{F}{t}{n})$ is computed using a small 3-layer convolutional sub-network.
The per-pixel weight of each view is obtained by applying softmax to the scores: $A_{t,n} = \mathrm{exp}(E_{t,n})/\sum_{m \in N} \mathrm{exp}(E_{t,m})$.
Finally, the aggregation function becomes $\sum_{n \in N} A_{t,n} \cdot \warp{F}{t}{n}$.
In both cases, pixels deemed occluded are masked out.

In our model, we apply this warping and aggregation to every skip connection and to the encoder output, thus mixing information across views and scales.
For every view in a batch, we aggregate all the other overlapping views within the batch.

\input{figures/splits.tex}
%%%%%%%%%%%%%%%%%%%%%%%%%%%%%%%%%%%%%%%%%%%%%%%%%%%%%%%%%%%%%%%%%%%%%%%%%%%%%%%%
\subsubsection{Feature Space Reprojection}
When the network is trained to make predictions one input view at a time, the feature maps at intermediate layers within the network contain viewpoint-dependent features, as illustrated in Figure~\ref{fig:feature-aggregation}.
Warping a feature map from one view to another only aligns features using the scene geometry, but does not change their dependence on viewpoint.

Imagine, for example, two \textsc{rgb-d} images in an urban scene – one from above looking down, and one at the road level.
The surface of the road will have the same colour in both views.
The depth of the road, however, will be different:
in the top view, it will be mostly uniform, while in the road-level view it will increase towards the horizon.
We can easily establish correspondences between pixels in the two images if we know their relative pose.
When not occluded, corresponding pixels will have the same colour, but not necessarily the same depth.
This is because colour is view-point independent, while depth depends on the viewpoint.
We know, however, how to compute one of the depth values, given the corresponding pixel in the other view, and the relative pose.
Thus, armed with this observation we propose a way to learn a mapping of \emph{features} between viewpoints.

Consider again two views $t$ and $n$, and a warped feature map $\warp{F}{t}{n}$ from the vantage point $n$.
Given a spatial location $\mathbf{p}$, we have a feature vector $\mathbf{f}_n = \warp{F}{t}{n}(\mathbf{p}) \in \mathbb{R}^\mathrm{D}$, its corresponding location in space $\mathbf{x}_n = (x, y, z, w) \in \mathbb{P}^3$, and a transform $\transform{t}{n} \in \mathrm{SE}(3)$.
We compute a matrix $\mathbf{W} = g(\transform{t}{n}) \in \mathcal{M}_{\mathrm{D},\mathrm{D+4}}$, using a small \ac{mlp} to model $g$.
This allows us to learn a linear transform in \emph{feature space} between view $n$ and view $t$:

\begin{equation}
\mathbf{f}_t = \mathbf{W} \left[\begin{matrix}
\mathbf{f}_n \\
\mathbf{x}_n
\end{matrix}\right].
\end{equation}
Without this mechanism, the network would have to learn to extract viewpoint-independent features to allow for feature aggregation between views.

Concretely, we implement this as a \ac{dfn}, with a 4-layer \ac{mlp} generating filters for a $1\times1$ linear convolution of the warped feature map $\warp{F}{t}{n}$.
To keep the \ac{mlp} small, we first project the input feature maps to $\mathrm{D} = 32$ dimensions, apply the dynamically generated filters, and then project back to the desired number of features.
We use the same transformation $\mathbf{W}$ across each of the the scales that we aggregate.

In the experiments, we show that this mechanism is essential for enabling effective multi-view aggregation.

%%%%%%%%%%%%%%%%%%%%%%%%%%%%%%%%%%%%%%%%%%%%%%%%%%%%%%%%%%%%%%%%%%%%%%%%%%%%%%%%
\subsubsection{Losses}\label{sec:losses}
We supervise our training with labels from a high-quality reconstruction, as sown in Figure~\ref{fig:training}. 
The labels provide two per-pixel supervision signals, one for direct regression, $\loss{\dataname}$, and one for prediction gradients, $\loss{\gradname}$:

\begin{equation}\label{eqn:dataloss}
  \loss{\dataname} = %
    \sum_{\mathbf{p} \in V} \left\|\pred - \gtruth\right\|_{berHu}; 
\end{equation}%
\begin{equation}\label{eqn:gradloss}
   \loss{\gradname} = \frac{1}{2} \sum_{\mathbf{p} \in V} %
   \left(\left|\partial_x \pred{} - \partial_x \gtruth{} \right| %
    + \left|\partial_y \pred{} - \partial_y \gtruth{} \right| \right), %
\end{equation}%
where $V$ is the set of valid pixels (to account for missing data in the ground-truth), $\pred$ and $\gtruth$ are the prediction and the target, respectively, and $\|\cdot\|_{berHu}$ is the berHu norm \citep{Owen:2007:BerHu}, whose advantages for depth prediction have been explored by \cite{Laina:2016:Deeper,Ma:2018:SparseToDense}. 
We use the Sobel operator \citep{Sobel:1968} to approximate the gradients in Equation~\ref{eqn:gradloss}.

The geometric consistency loss guides nearby predictions to have the same \threed{} geometry, and relies on warped nearby views $\warp{\predidepth}{t}{n}$. 
For a target view $t$, a set of nearby views $N$, the set of pixels unoccluded in a nearby view $U_n$ (see Figure~\ref{fig:mesh-features}, top row), this loss is defined as:
\begin{equation}
   \loss{\gcname} = \sum_{n \in N}\sum_{p \in U_n} \left|\warp{\predidepth}{t}{n} - \warpreproj{\predidepth}{t}{n}\right|.
\end{equation}
Both $\warp{\predidepth}{t}{n}$ and $\warpreproj{\predidepth}{t}{n}$ are aligned with view $t$ and contain inverse-depth values in frame $n$, as per Equation~\ref{eqn:image-warp} and Equation~\ref{eqn:idepth-warp}.
Note that $U_n$ has no relation to the set of valid pixels ($V$) from the previous losses, since this loss is only computed between predictions.
This enables the network to make sensible predictions even in parts of the image which have no valid label.

Finally, we also include an $L_2$ weight regulariser, $\loss{\regname}$, to reduce overfitting by keeping the weights small. The overall objective is thus defined as:
\begin{equation}
    \loss{} = %
        \scaling{\dataname} \loss{\dataname} %
        + \scaling{\gradname} \loss{\gradname} %
        + \scaling{\gcname} \loss{\gcname} %
        + \scaling{\regname} \loss{\regname}, %
\end{equation}
where the $\lambda$s are weights for each of the components.
We use $\scaling{\dataname} = 1$, $\scaling{\gradname} = 0.1$, $\scaling{\gcname} = 0.1$, and $\scaling{\regname} = 10^{-6}$.

%% file: figures/mesh-features.tex
%%% Help commands for images
\newcommand{\mfeat}[2]{%
\includegraphics[trim={0 0.5cm 0 0.5cm},clip,width=0.23\textwidth,frame={\fboxrule} {-\fboxrule}]{%
figures/graphics/mesh-features/#1_#2.png}}
\newcommand{\mfview}[2]{
\subfigure[#2]{
\parbox{0.23\textwidth}{
\mfeat{occlusion}{#1}\\[0.6ex]
\mfeat{idepth}{#1}\\[0.6ex]
\mfeat{colour}{#1}\\[0.6ex]
\mfeat{normals}{#1}\\[0.6ex]
\mfeat{area}{#1}\\
\vspace{-1.5ex}
}}}%
%
%%%%% actual figure
\begin{figure*}[t!]
\centering
% \subfigure[Ground truth]{%
% \parbox{0.16\textwidth}{
% \labs{019}{0}\\[1.2ex]
% \labs{004}{0}\\[1.2ex]
% \labs{118}{3}
% }}\hfill
%
\mfview{0}{Left}\hfill
\mfview{1}{Right}\hfill
\mfview{2}{Back}\hfill
\mfview{3}{Top}
\caption{%
Example of training data generated.
Each column represents a different view rendered around the same location.
The top row shows the inverse-depth images rendered from the lidar reconstruction, with areas visible in the other views shaded: red for left, green for right, blue for back, and cyan for top.
The next four rows show the mesh features we render from the stereo camera reconstruction: inverse-depth, colour, normals, and triangle surface area.
Our proposed model learns to refine the low-quality inverse-depth (second row), using the rendered mesh features (rows 2–5) as input, processing all four views jointly, and supervised by the high-quality inverse-depth label (first row).
}
\vspace{-2ex}
\label{fig:mesh-features}
\end{figure*}

%% file: figures/training.tex
\begin{figure*}[t!]
\centering
\includegraphics[width=\textwidth]{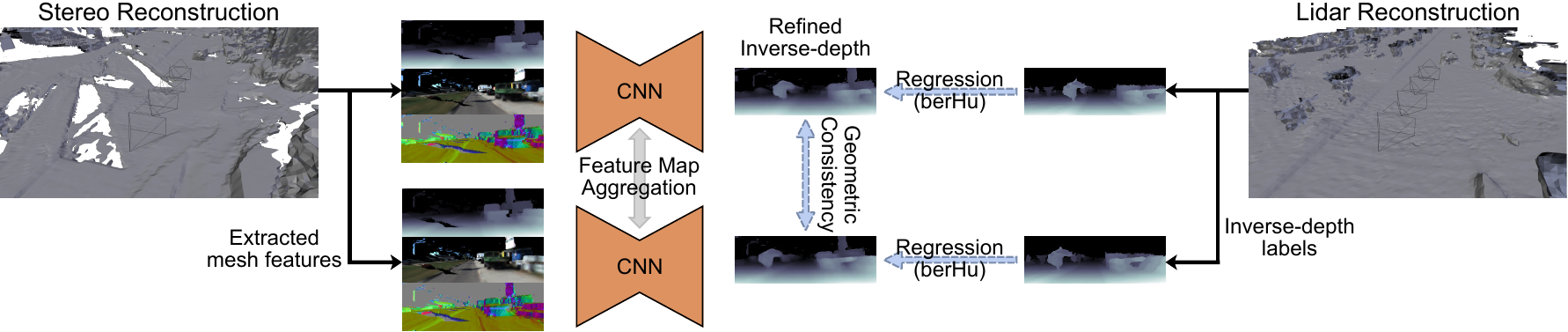}
\caption{
Illustration of our training set-up.
Starting with a low-quality reconstruction (stereo camera in this instance), we extract mesh features from several viewpoints.
Our network learns to refine the input inverse-depth with supervision from a high-quality (lidar) reconstruction.
The blue arrows above indicate the losses used during training: for each view, we regress to the high-quality inverse-depth, as well as to its gradient;
between nearby prediction, we apply a geometric consistency loss to encourage predictions with the same geometry.
Within the network, feature maps are aggregated between views so information can propagate within a neighbourhood of views.
}
\vspace{-1.5ex}
\label{fig:training}
\end{figure*}

%% file: tables/network-architecture.tex
\begin{table}[t!]
    \centering
    \caption{Overview of the \acs{cnn} architecture for error prediction}
    \begin{tabular}{lcc}
        \toprule
        \textbf{Block Type} & \textbf{Filter Size/Stride} & \textbf{Output Size} \\
        \midrule
        Input & - & $96\times 288\times \phantom{00}\textrm{F}$ \\
        Projection & $7\times 7$/$1$ & $96\times 288\times \phantom{0}16$ \\
        Residual & $3\times 3$/$1$ & $96\times 288\times \phantom{0}16$ \\
        Projection, Residual$\times2$ & $3\times 3$/$2$ & $48\times 144\times \phantom{0}32$ \\
        Projection, Residual$\times2$ & $3\times 3$/$2$ & $24\times \phantom{0}72\times \phantom{0}64$ \\
        Projection, Residual$\times2$ & $3\times 3$/$2$ & $12\times \phantom{0}36\times 128$ \\
        Projection, Residual$\times5$ & $3\times 3$/$2$ & $\phantom{0}6\times \phantom{0}18\times 256$ \\
        Up-projection, Skip & $3\times 3$/$\frac{1}{2}$ & $12\times \phantom{0}36\times 384$ \\
        Up-projection, Skip & $3\times 3$/$\frac{1}{2}$ & $24\times \phantom{0}72\times 192$ \\
        Up-projection, Skip & $3\times 3$/$\frac{1}{2}$ & $48\times 144\times \phantom{0}96$ \\
        Up-projection, Skip & $3\times 3$/$\frac{1}{2}$ & $96\times 288\times \phantom{0}48$ \\
        Residual$\times2$ & $3\times 3$/$1$ & $96\times 288\times \phantom{0}48$ \\
        Convolution & $3\times 3$/$1$ & $96\times 288\times \phantom{00}1$ \\
        \bottomrule
    \end{tabular}
\label{tab:network-architecture}
\end{table}

%% file: figures/splits.tex
\begin{figure*}[h]
\centering
\subfigure[Sequence 00]{\includegraphics[width=0.31\textwidth]{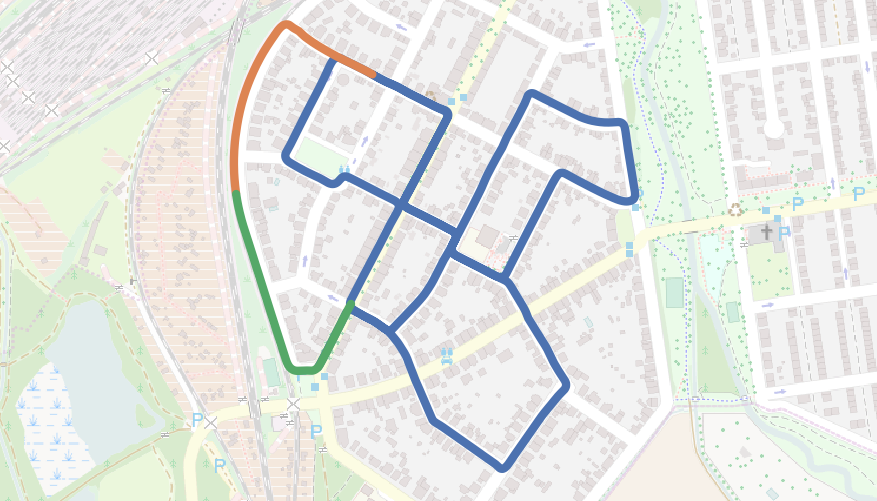}}%
\hfill%
\subfigure[Sequence 05]{\includegraphics[width=0.31\textwidth]{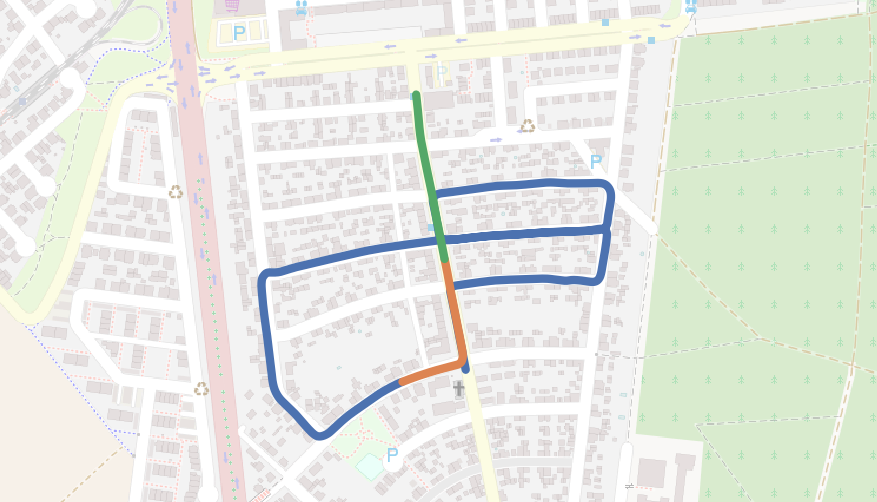}}%
\hfill%
\subfigure[Sequence 06]{\includegraphics[width=0.31\textwidth]{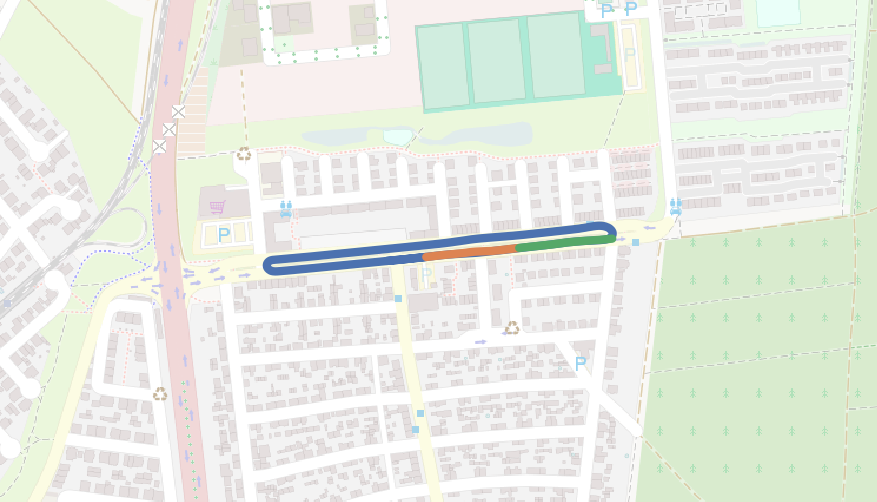}}
\caption{
Illustration of the training (blue), validation (orange), and test (green) splits on the three \acs{kitti-vo} sequences we are using.
Map data copyrighted OpenStreetMap \cite{OpenStreetMap} contributors and available from \url{https://www.openstreetmap.org}.
}
\vspace{-1ex}
\label{fig:splits}
\end{figure*}

%% file: sections/4_experiments.tex
\section{Experiments}
\label{sec:experiments}

\subsection{Experimental Setup}
\input{tables/results-main.tex}
\input{tables/results-generalisation.tex}
\subsubsection{Dataset}
For the experiments, we use sequences “00”, “05” and “06” from the \ac{kitti-vo} dataset \citep{Geiger:2013:KITTI}.
Using the \borg{} reconstruction system \citep{Tanner:2015:BORG}, we create pairs of low/high quality reconstructions (meshes) from the stereo camera, and lidar, respectively.
Following the same trajectory used when collecting data (as it is collision-free), every 0.65\,m we render mesh features from four views (left, right, back, top), illustrated in \cref{fig:mesh-features}.
For each view, we render a further 3 samples with small pose perturbations for data augmentation.
In total, we obtain 178\,544 distinct views of size $96 \times 288$ over 7.2\,km.

\subsubsection{Training and inference}
We train all our models on Nvidia~Titan~V \textsc{gpu}s, using the Adam optimiser \cite{Kingma:2015:Adam}, with $\beta_1 = 0.9$, $\beta_2 = 0.999$, and a learning rate that decays linearly from $10^{-4}$ to $5 \cdot 10^{-6}$ over 120\,000 training steps.
We clip the gradient norm to 80.
Each training batch contains 4 different examples, and each example is composed of the four views rendered around a single location.
Unless otherwise mentioned, we train our models for 500\,000 steps.
During inference, our full model runs at 11.3\,Hz when aggregating 4 input views, compared to the baseline that runs at 12.6\,Hz, so our method comes with little computational overhead.

\subsubsection{Metrics}
As our method operates on \twod{} views extracted from the mesh we are correcting, we measure how well our network predicts inverse-depth images, with the idea that better inverse-depth images result in better reconstructions.
We employ several metrics common in the related tasks of depth prediction and refinement.

One way to quantify performance is to see how often the error in prediction is small enough to be correct.
The thresholded accuracy measure is essentially the expectation that a given pixel is within a fraction $thr$ of the label:
\begin{equation}\label{eq:thresholded_accuracy}
    \delta = \mathbb{E}_{p \in V} \left[
        \mathbb{I}\left(\textrm{max}\left(
            \frac{\hqidepth}{\predidepth}, 
            \frac{\predidepth}{\hqidepth}\right) 
        < thr\right)
    \right],
\end{equation}
where $\hqidepth$ is the reference inverse-depth, $\predidepth$ is the predicted inverse depth, $V$ is the set of valid pixels, $n$ is the cardinality of $V$, and $\mathbb{I(\cdot)}$ represents the indicator function. For granularity, we use $thr = \in \{1.05, 1.15, 1.25, 1.25^2, 1.25^3\}$. 

In addition, we also compute the \ac{mae} and \ac{rmse} metrics to quantify per pixel error:
\begin{align}
\textrm{iMAE} &= \frac{1}{n}\sum_{p \in V}{\left|\predidepth - \hqidepth\right|}, \\
\textrm{iRMSE} &= \sqrt{\frac{1}{n}\sum_{p \in V}{(\predidepth - \hqidepth)^2}},
\end{align}
where the ‘i’ indicates that the metrics are computed over inverse-depth images.

\subsection{Gross Error Correction}
For the first set of experiments, we take the first 80\% of the views from each sequence as training data, the next 10\% for validation, and show our results on the last 10\%.
An illustration of the \textsc{kitti} sequences and splits is shown in \cref{fig:splits}.

As baseline, we train our model with geometric consistency loss but without any feature aggregation.
During inference, this model makes predictions one view at a time.

To illustrate our method, we train a further two models for each aggregation method (averaging and attention): one with the feature transform disabled, and one with it enabled.

As it can be seen in \cref{tab:results-main}, the baseline already refines inverse-depth significantly.
Without our feature transformation, the models are unable to use multi-view information because of the vastly different viewpoints, and indeed this slightly hurts performance.
Only when transforming the features between viewpoints does the performance increase over the baseline, highlighting the importance of our method for successfully aggregating multiple views.

\subsection{Generalisation}
To asses the ability of our method to generalise on unseen reconstructions, we divide our training data by sequence: we use two of the sequences for training, and the third for testing.
Sequences 00 and 05 are recorded in a suburban area with narrow roads, while sequence 06 is a loop on a divided road with a median strip, a much wider space and visually distinct.
We train models for 200\,000 steps and aggregate feature maps by averaging.
The results in \cref{tab:results-generalisation} show that our method successfully uses information from multiple views, even in areas of a city different from the ones it was trained on.
Furthermore, they reaffirm the need for our feature transformation method in addition to warping.

%% file: tables/results-main.tex
\begin{table*}
\centering
\caption{Depth Error Correction Results}
\label{tab:results-main}
\pgfplotstableread{tables/data/main-results.csv}\data
\pgfplotstabletypeset[
    highlight col max ={\data}{thacc1.05}, highlight col secondmax ={\data}{thacc1.05},
    highlight col max ={\data}{thacc1.15}, highlight col secondmax ={\data}{thacc1.15},
    highlight col max ={\data}{thacc1.25}, highlight col secondmax ={\data}{thacc1.25},
    highlight col max ={\data}{thacc1.56}, highlight col secondmax ={\data}{thacc1.56},
    highlight col max ={\data}{thacc1.95}, highlight col secondmax ={\data}{thacc1.95},
    highlight col min ={\data}{imae}, highlight col secondmin ={\data}{imae},
    highlight col min ={\data}{irmse}, highlight col secondmin ={\data}{irmse},
    string replace={0}{},
    columns/Model/.style={ column type={l|}, }, 
    every row no 1/.style={ after row=\midrule, },
    every row no 3/.style={ after row=\midrule, },
]{\data}
\end{table*}

%% file: tables/results-generalisation.tex
\begin{table*}[h]
\centering
\caption{Generalisation Capability of Depth Error Correction}
\label{tab:results-generalisation}
\begin{filecontents*}{tables/data/generalisation.csv}
Model, Train, Test, imae, irmse, thacc1.05, thacc1.15, thacc1.25, thacc1.56, thacc1.95
Uncorrected,                      –, 06, 0.02429,0.08391,0.49631,0.72893,0.75075,0.78481,0.80233
  Baseline,                {00; 05}, 06, 0.02091,0.08206,0.59987,0.79158,0.82623,0.86330,0.88544
Ours (no feat. transform), {00; 05}, 06, 0.02057,0.08108,0.61571,0.77951,0.81461,0.85450,0.87877
Ours (w/ feat. transform), {00; 05}, 06, 0.01943,0.06804,0.63901,0.78796,0.82003,0.85805,0.88035
#
Uncorrected,                      –, 05, 0.01154,0.03589,0.55845,0.81596,0.85074,0.88857,0.90898
  Baseline,                {00; 06}, 05, 0.00776,0.02979,0.72233,0.87542,0.91229,0.95458,0.97411
Ours (no feat. transform), {00; 06}, 05, 0.00779,0.02970,0.71574,0.87200,0.90938,0.95313,0.97337
Ours (w/ feat. transform), {00; 06}, 05, 0.00765,0.02956,0.73238,0.87607,0.91163,0.95338,0.97293
#
Uncorrected,                      –, 00, 0.01188,0.03240,0.53532,0.80456,0.85177,0.89554,0.91380
  Baseline,                {05; 06}, 00, 0.00840,0.02499,0.67516,0.86830,0.91283,0.95845,0.97701
Ours (no feat. transform), {05; 06}, 00, 0.00832,0.02485,0.66037,0.86651,0.91341,0.95805,0.97643
Ours (w/ feat. transform), {05; 06}, 00, 0.00840,0.02531,0.67895,0.87016,0.91388,0.95830,0.97643
\end{filecontents*}
\pgfplotstabletypeset[
    % columns={Model,Train,Test,thacc1.05,thacc1.25,thacc1.95},
    columns/Model/.style={ column type={l}, },
    columns/Test/.style={ column type={c|}, },
    every row no 3/.style={after row=\midrule},
    every row no 7/.style={after row=\midrule},
]{tables/data/generalisation.csv}
\end{table*}

%% file: sections/6_conclusion.tex
\section{Conclusion and Future Work}
\label{sec:conclusion}
In conclusion, we have presented a new method for correcting dense \threed{} reconstructions via \twod{} \emph{mesh feature} renderings.
In contrast to previous work, we make predictions on multiple views at the same time by warping and aggregating feature maps inside a \ac{cnn}.
In addition to warping the feature maps, we also transform the \emph{features} between views and show that this is necessary for using arbitrary viewpoints.

The method presented here aggregates feature maps between every pair of overlapping input views.
This scales quadratically with the number of views and thus limits the size of the neighbourhood we can reasonably process.
Future work will consider aggregation into a shared \twod{} spatial representation, such as a 360° view, which would scale linearly with the input neighbourhood size.